\documentclass{article}
\usepackage{ijcai19}

\usepackage{times}
\usepackage{soul}
\usepackage{url}
\usepackage[hidelinks]{hyperref}
\usepackage[utf8]{inputenc}
\usepackage[small]{caption}
\usepackage{graphicx}
\graphicspath{{fig/}}
\usepackage{amsmath}
\usepackage{amsfonts}
\usepackage{booktabs}
\usepackage{algorithm}
\usepackage{algorithmic}
\usepackage{cancel}
\usepackage{multicol}
\usepackage{multirow}
\usepackage{xcolor}
\usepackage{caption}

\usepackage[round]{natbib}

\title{Diversity-Inducing Policy Gradient:\\
           Using Maximum Mean Discrepancy to Find a Set of Diverse Policies}

\author{
Muhammad A Masood$^1$\footnote{Contact Author}\and
Finale Doshi-Velez$^1$\\
\affiliations
$^1$Harvard University\\
\emails
masood@g.harvard.edu,
finale@seas.harvard.edu
}

\begin{document}

\maketitle

\begin{abstract}
 Standard reinforcement learning methods aim to master one way of solving a task whereas there may exist multiple near-optimal policies. Being able to identify this \emph{collection} of near-optimal policies can allow a domain expert to efficiently explore the space of reasonable solutions.  Unfortunately, existing approaches that quantify uncertainty over policies are not ultimately relevant to finding policies with qualitatively distinct behaviors.  In this work, we formalize the difference between policies as a difference between the distribution of trajectories induced by each policy, which encourages diversity with respect to both state visitation and action choices.  We derive a gradient-based optimization technique that can be combined with existing policy gradient methods to now identify diverse collections of well-performing policies.  We demonstrate our approach on benchmarks and a healthcare task.
\end{abstract}

\section{Introduction}
\label{sec:intro}

Standard reinforcement learning methods find one way to solve a task, even though there may exist multiple near-optimal policies that are distinct in some meaningful way.  Identifying this \emph{collection} of near-optimal policies can allow a domain expert to efficiently explore the space of reasonable solutions.  For example, knowing that there exist comparably-performing policies that trade between several small doses of a drug or a single large dose may enable a clinician to identify what might work best for a patient (e.g., based on whether the patient will remember to take all the small doses).

Unfortunately, existing approaches to find a set of diverse policies involve notions of diversity that are not aligned with the kind of efficient exploration-amongst-reasonable-options setting described above.  \citet{liu2017stein} characterize the uncertainty over policies via computing a posterior over policy parameters, but differences in policy parameters may not result in qualitatively different behavior (especially in over-parameterized architectures).  \citet{haarnoja2017reinforcement} encourage diversity via encouraging high entropy distributions over actions (given states), which may result in sub-optimal behavior.  \citet{fard2011non} seek a single non-deterministic policy that may make multiple decisions at any state, which may be overly restrictive if action choices across states must be correlated to achieve near-optimal performance. 

We argue that differences in trajectories (state visits and action choices) better capture the kinds of distinct behavior we are seeking.  For example, does one prefer a policy that achieves wellness through a surgery, or via prolonged therapy?  More formally, stochasticity in the environment dynamics and the policy will induce a distribution over trajectories.  We use the maximum mean discrepancy (MMD) metric to compare these distributions over trajectories under different policies.  As noted in \citet{sriperumbudur2010non}, the MMD metric has a closed form solution (unlike the Wasserstein and Dudley metrics) and exhibits better convergence behavior than $\phi$-divergences such as Kullback-Leibler (KL).  

In this work, we first formalize notions of policy diversity via the MMD over their induced trajectories.  We show that unbiased gradient estimates of the MMD term can be obtained without knowledge of transition dynamics of the environment, and describe how it can be applied to \emph{any} policy gradient objective. Across both benchmark domains and a healthcare task, our approach discovers diverse collections of well-performing policies.

%\begin{figure}[t]
%  \centering
%    \includegraphics[width=0.35\textwidth]{barrier_demo.png}
%  \caption{After specifying one path to the goal (black line), we run our algorithm to find a policy that reaches the goal by going around the barrier in a different way (blue dots). Note that our algorithm reaches the goal state in a distinct and near-optimal manner.}
%    \label{fig:barrier_demo}
%\end{figure}

\section{Background}

\paragraph{Reinforcement Learning}
We consider Markov decision processes (MDPs) defined by a continuous state space $\mathcal{S}$, a (discrete or continuous) action space $\mathcal{A}$, state transition probabilities $p_T(s,a,s')$, reward function $r(s,a)$, as well as a discount factor $\gamma$.  A policy $\pi(s,a)$ indicates the probability of action $a$ in state $s$; together with the transition probability $p_T(s,a,s')$, it induces a distribution $p_{\pi}(\tau)$ over trajectories $\tau = s_0, a_0, \hdots a_{T-1},s_T$.  Traditionally, the task is to find a policy that maximizes the long-term expected discounted sum of rewards (return) $g(\tau)$:
\begin{equation*}
g(\tau) = \mathbb{E}_{{\tau \sim p_{\pi}(\tau)}} \big[ \sum_{t=0}^{T} \gamma^t r(s_t, a_t) \big]
\end{equation*}
In this work, we shall seek multiple near-optimal policies.  

\paragraph{Policy Gradient Methods}
Policy gradient methods use gradients to iteratively optimize a policy $\pi_\theta(s,a)$ that is parameterized by $\theta$.  The standard objective $J_\text{PG}(\theta)$ is the return $g(\tau)$.  An unbiased estimate of the gradient of the objective can be obtained by Monte Carlo rollouts generated by the policy $\pi_\theta$ using the likelihood ratio trick. For a single rollout $\{s_t, a_t, r_t\}$, the gradient can be estimated as
\begin{equation*}
\nabla_\theta J_\text{PG}(\theta) \approx \sum_{t = 0}^{T} \nabla_\theta \log \pi_{\theta}(s_t, a_t) g_t
\end{equation*}
where $g_t = \sum_{t' = t}^{T} \gamma^{t' - t}r_{t'}$ is the return over the rewards received from time $t$ onwards.
\citet{schulman2017proximal} introduced Proximal Policy Optimization (PPO); a popular variant that achieves state-of-the-art performance on many benchmark domains and uses trust region updates that are compatible with stochastic gradient descent. 

In this work we augment the PPO objective for the policy network with an MMD-based term which encourages policies that lead to different state visitation and/or action choice distribution than previously-identified policies. An iterative use of this diversity inducing variant of policy gradient methods allows us to sequentially obtain distinct policies.  

\paragraph{Off-Policy Evaluation}
The off-policy RL framework applies when trajectories are collected from a behavior policy that is distinct from the policy being trained, e.g., using observations from clinician behavior to optimize an agent's behavior. In the batch setting, there are model-based and importance sampling approaches to estimating the value of a policy. Model-based approaches learn a dynamics model for the transitions and use those to simulate outcomes of a policy in order to estimate its value. Importance sampling approaches appropriately re-weight existing batch data to estimate the value for a different policy. %In this work, we use an importance sampling estimator (Consistent Weighted Per-Decision Importance Sampling - CWPDIS) proposed in \citet{thomas2015safe}. 

% \citet{thomas2016data}

%Evaluating the value of a policy when only batch data is available requires either model-based or importance sampling approaches

%To compute an off-policy policy gradient, \citet{degris2012off} propose applying importance weights to the policy gradient. We will take advantage of this work when finding diverse policies in ICU settings where experimentation is not possible. 
% Think about other references from Guided Policy Search Paper (Levine) 
% ; their derivation assumes that the gradient of the state-action value function of the target policy with respect to the policy parameters can be ignored.
%More recently, \citet{imani2018off} establish an off-policy policy gradient theorem and develop an actor-critic algorithm for learning in the off-policy setting.  

\paragraph{Maximum Mean Discrepancy}
We use the MMD metric to measure the difference between two trajectory distributions. The MMD is an integral probability metric \citep{gretton2007kernel} that measures the difference between two distributions $p, q$ using test functions $h$ from a function space  $\mathcal{H}$. It is given by: 
\begin{equation*}
\text{MMD}(p,q,\mathcal H) = \sup_{h \in \mathcal H} (\mathbb{E}_{x \sim p}[h(x)] -  \mathbb{E}_{y \sim q}[h(y)] )
\end{equation*}
Computing the MMD is tractable when the function space $\mathcal{H}$ is a unit-ball in a reproducing kernel hilbert space (RKHS) defined by a kernel $k(\cdot,\cdot)$ and is given by:
\begin{equation*}
\text{MMD}^2(p,q,\mathcal H) = \mathbb{E} [ k(x,x') ] -2 \mathbb{E} [ k(x,y) ]  +  \mathbb{E} [ k(y,y') ] 
\end{equation*}
where $x, x' \enskip \text{i.i.d.} \sim p$ and $y,y' \enskip \text{i.i.d.} \sim q$. The expectation terms in the analytical expression for the MMD can be approximated using samples.  In Section~\ref{sec:method}, we will show that computing the derivative of the MMD metric between trajectories with respect to the policy parameters $\theta$ is tractable.

\section{Diversity-Inducing Policy Gradient (DIPG)}
\label{sec:method}

Our algorithm constructs a set of diverse policies for an MDP by iteratively finding policies that are diverse relative to an existing set of policies. First, we formulate a diversity inducing objective function that regularizes any policy gradient objective. Then, we show that optimizing this objective is tractable using the familiar log-derivative trick. Next, we explain how to iteratively apply this diversity-inducing policy gradient objective in conjunction with an existing algorithm to find a set of distinct policies that solve an MDP. Finally, we introduce an extension to the proposed framework for off-policy batch reinforcement learning. 

\subsection{DIPG Objective}
We propose adding a regularization term to encourage learning a policy $\pi_\theta(s,a)$ that induces a distribution over trajectories $p_{\theta}(\tau)$ that is distinct from a specified set of distributions over trajectories $\mathcal{Q} = \{q_{m}(\tau)\}_{m = 1}^M$.  (Below, these distributions $q_m(\tau)$ will come from previously-identified policies.)  Our diversity measure $D_{\text{MMD}}(p_{\theta}(\tau),\mathcal{Q})$ is the squared MMD between the distribution of trajectories under the current policy $\pi_\theta$ and the distribution $q_{*}(\tau)$ in $\mathcal{Q}$ most similar to it.
\begin{equation*}
\begin{split}
D_{\text{MMD}}(p_{\theta}(\tau),\mathcal{Q}) &= \min_{m \in \{1,\hdots,M\}} \text{MMD}^2(p_{\theta}(\tau),q_{m}(\tau)) \\
&= \text{MMD}^2(p_{\theta}(\tau),q_{*}(\tau))
\end{split}
\end{equation*}

We define a kernel over a pair of trajectories $(\tau, \tau')$ using a kernel $K$ (such as Gaussian kernel) defined over some function ($g$) of a vectorized representation of $N$ steps of trajectories ($x = \phi_N(\tau), x' = \phi_N(\tau')$). For trajectories that are not of the same length, we pick $N$ to be the number of steps corresponding to the shorter of the two trajectories. 

\begin{equation*}
k(\tau, \tau') = K(g(\phi_N(\tau)), g(\phi_N(\tau')))
\end{equation*}

The function $\phi_N$ is simply a way to stack the states and actions from the first $N$ steps of a trajectory into a single vector. The function $g$ gives the flexibility to adjust the focus of the diversity measure for different aspects such as state visits, action choices, or both. 
% FDV: Can you mention you'll give examples of g later / write out in the supplement? 

The MMD-based diversity-inducing objective function is given by $J_{\text{MMD}}(\theta)$:
\begin{equation*}
J_{\text{MMD}}(\theta) = J_{\text{PG}}(\theta) + \alpha D_{\text{MMD}}(p_{\theta}(\tau),\mathcal{Q})
\end{equation*}

Where $J_{\text{PG}}(\theta)$ is the objective function of a policy gradient algorithm of the user's choice (e.g. for vanilla policy gradient it would be the expected return) and the second term is proportional to the diversity measure between a policy distribution and the set of specified distributions $\mathcal{Q}$. The parameter $\alpha$ decides the relative importance of optimality and diversity of the new policy $\pi_\theta(s,a)$. 

To optimize the MMD-based diversity-inducing objective, we need to specify how gradients of the diversity term can be computed with respect to the policy parameters $\theta$. 

\subsection{Optimization via Gradient Ascent}
To use gradient ascent on  $J_{\text{MMD}}(\theta)$, what remains to specify is the gradient with respect to $\theta$ of the diversity inducing term $D_{\text{MMD}}$. Let $q_{*}(\tau)$ be the distribution in $\mathcal{Q}$ that minimizes the MMD between the state trajectory distribution of the policy $\pi_\theta$ and $q_m \in \mathcal{Q}$. Then, the gradient with respect to the policy parameters $\theta$ of the diversity term is given by
\begin{equation*}
\begin{split}
\nabla_\theta D_{\text{MMD}}(p_{\theta}(\tau),\mathcal{Q}) &= \nabla_\theta \text{MMD}^2(p_{\theta}(\tau),q_{*}(\tau),\mathcal H) \\
 &= \mathbb{E} [k(\tau_p,\tau_p')\nabla_\theta \log (p_\theta(\tau_p)p_\theta(\tau_p'))] \\ &-2\mathbb{E} [k(\tau_p,\tau_q)\nabla_\theta \log (p_\theta(\tau_p)q_{*}(\tau_q))]  \\ &+  \mathbb{E} [k(\tau_q,\tau_q')\cancelto{0}{\nabla_\theta \log (q_{*}(\tau_q)q_{*}(\tau_q'))}]  \\
\end{split}
\end{equation*}
where $\tau_p, \tau_p' \enskip \text{i.i.d.} \sim p_{\theta}(\tau)$ and $\tau_q,\tau_q' \enskip \text{i.i.d.} \sim q_{*}(\tau)$. The last term only involves the distribution $q_{*}(\tau) \in \mathcal{Q}$ that has no dependence on $\theta$. The gradient term can be estimated by linear combinations of the $\nabla_\theta \log p_\theta(\tau)$ involving the kernel between sample trajectories from the policy $\pi_\theta$ and with a set of specified trajectories from $\mathcal{Q}$. It is well-known \citep{sutton2000policy} that the gradient of the score function of the trajectory distribution $\nabla_\theta \log p_\theta(\tau_p)$ does not require the dynamics model and can be expressed in terms of the score function of the policy network ($\nabla_\theta \log p_{\theta}(\tau) =  \sum_{t = 0}^H \nabla_\theta \log \pi_\theta(a_t | s_t)$).

We now have all the machinery in place to augment \emph{any} existing policy gradient method with a diversity inducing term. Next, we will specify the basic algorithm for finding a policy that is diverse with respect to some specified set of trajectory distributions and then introduce an algorithm that leverages this to find a set of diverse policies. 

\begin{algorithm}[tb]
   \caption{MMD-based Diversity-Inducing Policy}
   \label{alg:div_policy}
\begin{algorithmic}
   \STATE {\bfseries Input:} Known policies $\mathcal{P}_{\text{known}}$, MDP $\{\mathcal{S},\mathcal{A}, p_s, r\}$, policy gradient objective $J_{\text{PG}}$, learning rate $\eta$, diversity parameter $\alpha$ 
   \STATE Initialize policy parameters $\theta$
   \STATE $\mathcal{Q} \leftarrow $ Sampled trajectories from policies in $\mathcal{P}_{\text{known}}$
   \REPEAT
   \STATE Generate an episode $s_0,a_0,r_1,\hdots,s_{T-1},a_{T-1},r_{T}$, following policy $\pi_{\theta}$
   \FOR {Each step $t = 0, \hdots, T - 1$}
   \STATE {\bfseries 1:} Estimate  $\nabla_\theta J_{\text{PG}}$ and $\nabla_\theta D_{\text{MMD}}(p_{\theta}(\tau),\mathcal{Q})$
    \STATE {\bfseries 2:} Update policy parameters via gradient ascent \\ $\theta \leftarrow \theta + \eta (\nabla_\theta J_{\text{PG}} + \alpha \nabla_\theta D_{\text{MMD}}(p_{\theta}(\tau),\mathcal{Q}))$
   \ENDFOR
   \UNTIL {convergence}
   \STATE {\bfseries Output:} policy $p_\theta$
\end{algorithmic}
\end{algorithm}

\begin{algorithm}[tb]
   \caption{DIPG}
   \label{alg:mmd_multi_policy}
\begin{algorithmic}
   \STATE {\bfseries Input:} Number of policies $N$, MDP $\{\mathcal{S},\mathcal{A}, p_s, r\}$, policy gradient objective $J_{\text{PG}}$, learning rate $\eta$, diversity parameters $\alpha_{1\hdots N}$ 
   \STATE Collection of known policies $\mathcal{P}_{\text{known}} = \emptyset$
   \FOR{$n=1$ {\bfseries to} $N$}
   \STATE {\bfseries 1:} Find a policy $p_n$ that is distinct from the current set of known policies  $\mathcal{P}_{\text{known}} $:
   \STATE $p_n \leftarrow$ Algorithm~\ref{alg:div_policy}$(\mathcal{P}_{\text{known}}, \text{MDP}, J_{\text{PG}},\eta,\alpha_n) $
   \STATE {\bfseries 2:} Add $p_n$ to the set of known policies $\mathcal{P}_{\text{known}}$:
   \STATE $\mathcal{P}_{\text{known}} \leftarrow \mathcal{P}_{\text{known}} \cup p_n$
   \ENDFOR
   \STATE {\bfseries Output:} Set of policies $\mathcal{P}_{\text{known}}$
\end{algorithmic}
\end{algorithm}

\subsection{Finding Multiple Diverse Policies}
Our goal is to find a collection of policies that perform optimally or near-optimally and are diverse in terms of the distributions over trajectories that they induce. In Algorithm~\ref{alg:div_policy}, we state how the diversity inducing term $D_{\text{MMD}}$  is used in order to learn a single policy that is distinct with respect to a specified set of distributions over trajectories $\mathcal{Q}$. In Algorithm~\ref{alg:mmd_multi_policy}, we iteratively apply Algorithm~\ref{alg:div_policy} to find the desired set of $N$ distinct policies (agents): The first policy is learned without any diversity term because there is no set of known policies to begin with. Subsequent policies are learned by initializing random policy parameters and training with the augmented objective function that encourages diversity with respect to previously discovered policies. The strength of the diversity parameter $\alpha$ can be varied (e.g., to seek diverse policies more aggressively as more policies are discovered).  

\subsection{DIPG Extension: Batch Off-Policy Setting}
Batch reinforcement learning \citep{lange2012batch} aims to learn policies from a fixed set of previously-collected trajectories.  It is common in domains such as medicine, dialogue management, and industrial plant control where logged data are plentiful but exploration is expensive or infeasible.

% Applications of reinforcement learning to observational health settings \citep{gottesman2019guidelines}, dialogue management \citep{pietquin2011sample} and other domains such as turbine control \citep{depeweg2016learning} where it is expensive or infeasible for an agent to be directly deployed in the environment necessitate the use of batch reinforcement learning techniques. 

We now extend our DIPG framework to this batch setting. In the on-policy case, we defined the DIPG diversity term as a kernel over the distribution of trajectories induced by different policies.  However, in the batch case, trajectories cannot be simulated from the learned policies.  Thus, we instead define the diversity as a kernel over the likelihoods of specific (observed) trajectories in the batch with respect to a policy.  

Specifically, let $\mathcal{T} = \{\tau_i\}_{i=1}^{I}$ be a batch of $I$ trajectories.  We use $p(\mathcal{T} | \pi) \in \mathbb{R}^{I}$ to indicate a vector where the $i$\textsuperscript{th} coordinate equals the probability of the $i$\textsuperscript{th} trajectory under the policy $\pi$ i.e. $p_i(\mathcal{T} | \pi) = p(\tau_i | \pi)$.  Now, the diversity term can be defined in an analogous fashion to the on-policy case where we compare our policy being optimized $\pi_\theta$ to the previous policies $\{q_1,...,q_M\}$:

\begin{equation*}
\begin{split}
D^{\mathcal{T}}_{\text{BATCH}}(\pi_\theta,\mathcal{Q}) &= \min_{m \in \{1,\hdots,M\}} k(p(\mathcal{T} | \pi_{\theta}),p(\mathcal{T} | q_{m}) \\
&=  k(p(\mathcal{T} | \pi_\theta),p(\mathcal{T} | q^*_{m}))
\end{split}
\end{equation*}

We also require a measure of quality for the policies.  \citet{levine2013guided} note that gradient-based optimization of importance sampling estimates is difficult with complex policies and long rollouts. We suggest a surrogate that is equal to the sum of the likelihoods of each trajectory in the batch $J_{\text{Surrogate}}^{\mathcal{T}}(\theta)  = \sum_{i = 1}^{I} p(\tau_i | \theta)$. This surrogate is more robust to optimize and encourages `safe' behavior from the agent, a desirable feature in the healthcare setting.

\begin{equation*}
J_{\text{BATCH}}(\theta) = J_{\text{Surrogate}}^{\mathcal{T}}(\theta) + \alpha D^\mathcal{T}_{\text{BATCH}}(\pi^\theta,\mathcal{Q})
% \label{eq:batch_obj}
\end{equation*}
While we optimize with respect to this surrogate, in our results, we still report on the value of the policy with respect to a standard importance sampling-based estimator (CWPDIS, from \citet{thomas2015safe}).

\section{Experimental Setup}

We augment Proximal Policy Optimization (PPO) \citep{schulman2017proximal} with the diversity inducing term (referred to as DIPG-PPO). We set the maximum number of steps taken by all baseline algorithms to be 1 million and set the maximum number of steps for each of the $N$ policies in DIPG-PPO to 0.2 million. Since we choose $N < 5$ for all our testing environments, the total number of steps to learn all $N$ policies is necessarily less than 1 million. In these experiments we set $\alpha_n = 1.0$ across all environments and iterations $n$\footnote{The performance of our algorithm may improve when this parameter is tuned based on the environment or which of the $N$ policies is being learned)}. For the MMD kernel, we choose the Gaussian kernel with bandwidth set to 1. Each algorithm is run 3 times to assess variability in performance, except for RR-PPO which does not have any inherent diversity component, so we run it with 10 random restarts to help it make up for this disadvantage.

\subsection{Environments}
\paragraph{Synthetic Environments}
We illustrate qualitative aspects of our approach on 2-dimensional navigational tasks.  

\textit{Multi-goal} environment of  \citet{haarnoja2017reinforcement} is solved by reaching one of four symmetrically placed goals in a continuous 2-D world. An ideal collection of diverse policies would include policies that solve the task via reaching different goals in different parts of the state space.

%\textit{Obstacle} environment involves overcoming a pit/barrier region to reach the goal. There exist two optimal policies (going around the barrier in two different ways) that are meaningfully distinct. This environment requires finding different policies to reach a single goal. 

\textit{Asymmetric goals} environment has one goal that is closer to the starting point and easier to reach than the second. While most agents find the region closest to the initial position, a distinct collection of policies solves the environment by also reaching the goal further away. We use this environment to explore the case where there exist a slightly sub-optimal solution that is clearly distinct from the optimal one. 

\paragraph{Benchmark Environments}

We evaluated the performance of our algorithm on standard benchmark environments:

\emph{Reacher}, \emph{Ant}, \emph{Humanoid} and \emph{Humanoid Flag Run}  \citep{schulman2017proximal,PyBullet-Gym}.

\paragraph{Clinical batch}

\textit{Hypotension} batch data is built from a cohort of patients from MIMIC critical care data set \citep{johnson2017mimic}. In this work, we aim to learn multiple treatment strategies using off-policy methods. Cohort and MDP formulation details are in the supplement\footnote{https://tinyurl.com/ijai2019-DIPG}
%In the MDP formulation of this domain, the agent can choose to give a discretized set of 36 actions (corresponding to all combinations of six categories of IV fluid and vasopressor dosage volumes). The reward is based on blood-pressure with linear penalties for blood pressure that is either higher or lower than the target range.% (and is always between 0 and 1). %We use a discount factor of 0.9. 
 % Reference Mike's paper here?

\subsection{Algorithm and Baselines}
We compare our approach [\textbf{DIPG-PPO}] to the following other algorithms that can encourage diverse behavior:

\textit{Random Restarts} [\textbf{RR-PPO}]: Independent runs of PPO without the diversity inducing regularization. Due to variability in initializations and experience, we obtain a collection of policies that are not exactly the same. 

\textit{Stein Variational Policy Gradient } [\textbf{SVPG}]: \citet{liu2017stein} use functional gradient descent to compute a point-based posterior distribution over the policy parameter space.  We report results for a collection of $8$ agents, and for comparison also on a single agent's performance. 

\textit{Deep Energy-Based Policy} [\textbf{Soft-Q}]: \citet{haarnoja2017reinforcement} present a way to train a single stochastic policy that is encouraged to have high entropy on the probability of the action (given state). The higher entropy encourages the agent to choose actions that are diverse. We try Soft-Q learning with the default entropy regularization parameter of $1$ as well as variants with 0.5 and 0. We choose the smaller (and zero) values of the regularization parameter to see the effect of the tradeoff between diversity and quality in this algorithm. %We use $16$ particles for the function gradient update for the action-space as was used in the original work.  

\section{Results}

We evaluate the performance based on both quality and similarity scores (figure~\ref{fig:main_results}). The quality is evaluated by averaging the returns coming from rollouts of the final stochastic policy (or collection of policies) and the similarity score comes from the kernel used to measure similarity between trajectories (smaller kernel values indicate higher diversity). Additional results can be found in our supplement\footnote{https://tinyurl.com/ijai2019-DIPG}.

\begin{figure*}
\begin{multicols}{4}
    \includegraphics[width=\linewidth]{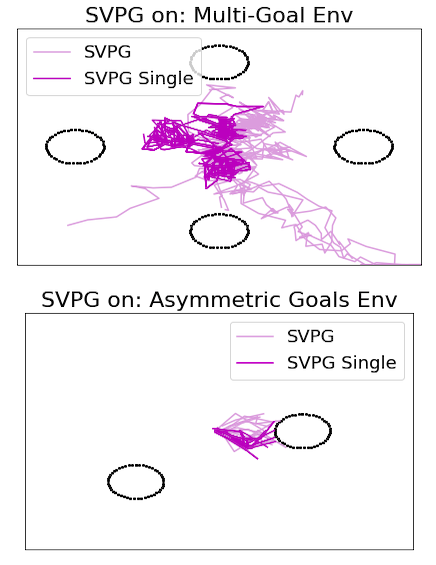}\par 
    \includegraphics[width=\linewidth]{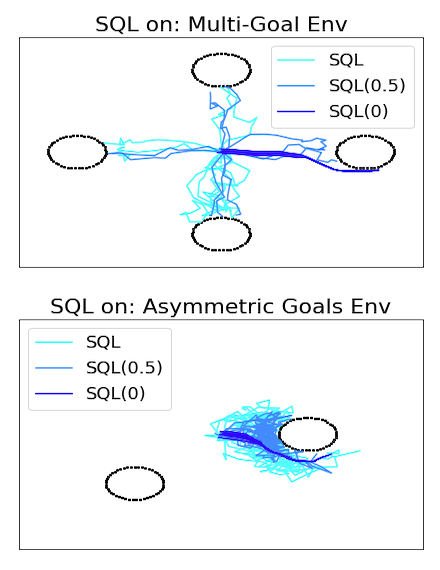}\par 
    \includegraphics[width=\linewidth]{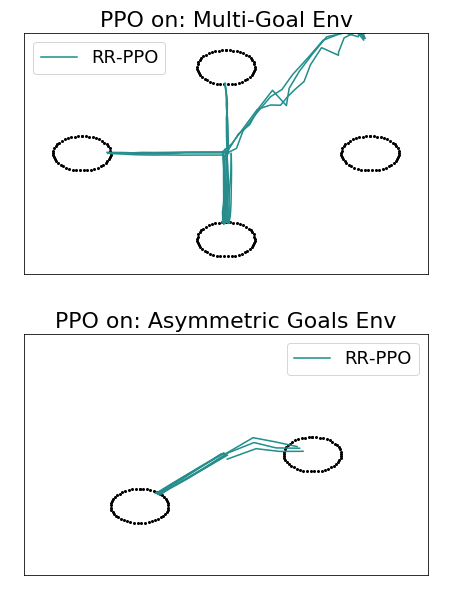}\par
    \includegraphics[width=\linewidth]{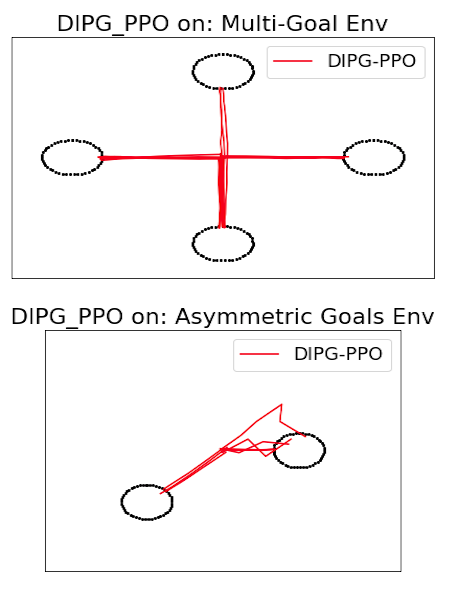}\par  
    \end{multicols}
\caption{A comparison of trajectories in the 2-D navigation tasks shows that DIPG-PPO (with $N = 4$ for Multi-Goal and $N = 2$ for Asymmetric Goals) produces near-optimal trajectories to reach each goal. Soft-Q takes sub-optimal paths to the goal regions and SVPG generally fails to solve the environments. }
\label{fig:quad_goal_trajs}
%\vspace{1pt}
\end{figure*}

\begin{figure*}
\begin{multicols}{3}
    \includegraphics[width=\linewidth]{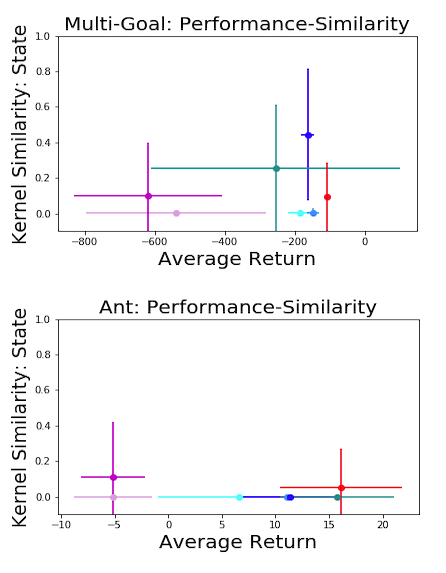}\par 
    \includegraphics[width=\linewidth]{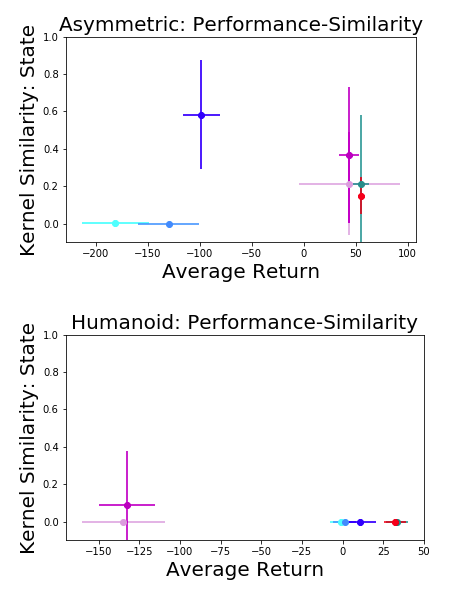}\par
    \includegraphics[width=\linewidth]{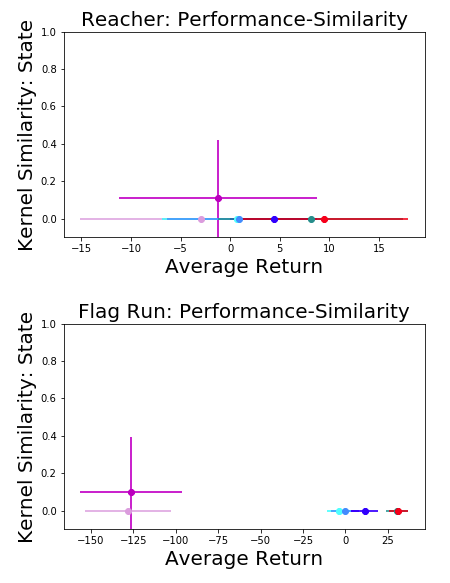}\par  
    \end{multicols}
\caption{A  comparison of the kernel-based similarity between trajectories and the average return across different algorithms reveals that DIPG-PPO and RR-PPO consistently give the largest returns while achieving low similarity. The colors are consistent with those in figure~\ref{fig:quad_goal_trajs}}
% FDV: If it's not too much trouble to add a legend in this figure, that would be great, but otherwise don't worry about it... 
\label{fig:main_results}
\end{figure*}

%\begin{figure*}
%\begin{multicols}{4}
%    \includegraphics[width=\linewidth]{Reacher_Return_Action.png}\par 
%    \includegraphics[width=\linewidth]{Ant_Return_Action.png}\par 
%    \includegraphics[width=\linewidth]{Humanoid_Return_Action.png}\par
%    \includegraphics[width=\linewidth]{FlagRun_Return_Action.png}\par  
%    \end{multicols}
%\caption{A quantitative comparison of the kernel-based similarity between trajectories and the average return across different algorithms reveals that DIPG-PPO and PPO random restarts consistently give the largest returns while achieving low similarity. Variants of Soft-Q with the entropy regularization parameter set to 0 and 0.5 are also shown as well as the performance of a single SVPG agent}
%%\label{fig:asymmetric_trajs}
%\end{figure*}

\begin{figure*}
\begin{multicols}{4}
    \includegraphics[width=\linewidth]{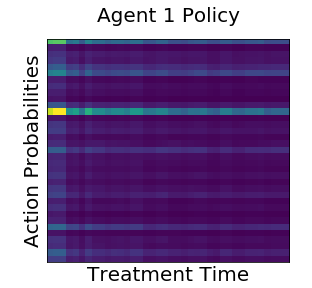}\par 
       \includegraphics[width=\linewidth]{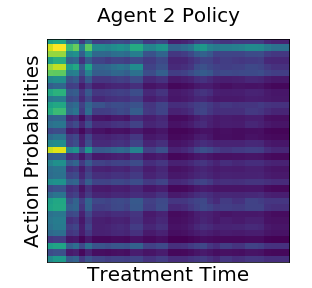}\par 
          \includegraphics[width=\linewidth]{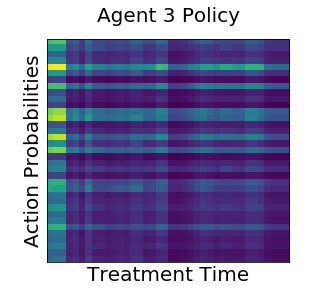}\par 
             \includegraphics[width=\linewidth]{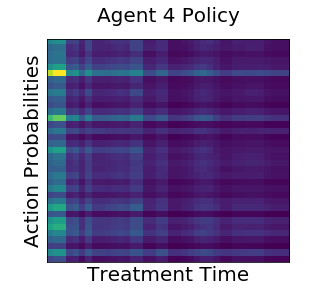}\par 
    \end{multicols}
\caption{We show the probability of taking actions throughout the duration of a single ICU stay.  While these agents all have roughly equal value, they exhibit different emphasis on treatment styles as can be seen by the variability in action probabilities.}
\label{fig:off-policy-results}
\end{figure*}

%\begin{figure*}
%\begin{multicols}{4}
%    \includegraphics[width=\linewidth]{Agent1policyT2.png}\par 
%       \includegraphics[width=\linewidth]{Agent2policyT2.png}\par 
%          \includegraphics[width=\linewidth]{Agent3policyT2.png}\par 
%             \includegraphics[width=\linewidth]{Agent4policyT2.png}\par 
%%    \includegraphics[width=\linewidth]{placeholder.png}\par 
%
%%    \includegraphics[width=\linewidth]{placeholder.png}\par 
% %   \includegraphics[width=\linewidth]{placeholder.png}\par  
%    \end{multicols}
%\caption{Space for off-policy results }
%\label{fig:off-policy-results}
%\end{figure*}

\paragraph{DIPG method does not sacrifice quality}
Figure~\ref{fig:main_results} shows that DIPG-PPO and RR-PPO generally provide the highest quality policies. The baseline algorithms designed for finding a diverse set of policies (SVPG and Soft-Q) have significantly worse quality performance (SVPG in particular) and find policies that lead to high variability in the rewards even though each individual policy in SVPG and Soft-Q is trained for longer (1 million steps) as compared to RR-PPO and DIPG-PPO (0.2 million steps). Such variability in the distribution of returns suggests that the diversity in SVPG and Soft-Q comes not only at the cost of quality, but importantly (especially in the clinical setting) at the cost of consistency of quality returns. It must be emphasized that diversity in policies is only meaningful if they are beyond some threshold of quality and are able to `solve' a domain consistently. 

\paragraph{DIPG method finds meaningfully diverse policies with fewer runs}
The diversity information in figure~\ref{fig:main_results} shows that SVPG and Soft-Q algorithms give policies that are quite diverse (comparison of pairwise trajectories shows negligible overlap). Since these collections of stochastic policies fail to provide high rewards in the environments, the diversity in the trajectory distribution that is induced is of little value. The DIPG-PPO ($N = 2$ and $N = 4$) and RR-PPO collections ($N = 10$) are collections of policies that essentially `solve' these environments and discover the set of ways these environments were designed to be solved (see figure~\ref{fig:quad_goal_trajs}). In comparison to RR-PPO, DIPG-PPO requires many fewer agents to discover an appropriate collection of diverse agents. 

\paragraph{Clinical Example: A collection of DIPG policies reveals different choices of treatment in the ICU.}
We learned four policies through our off-policy extension of DIPG using real data from the ICU and evaluated their performance using the CWPDIS importance sampling estimator \citep{thomas2015safe}. Not only were all four learned policies of slightly higher quality ($\sim9.33$ on held out test data) than the clinician behavior policy ($9.29$), they are also distinct %(average pairwise Frobenius distance between trajectory likelihood $\sim1.02$.

Figure~\ref{fig:off-policy-results} plots the probabilities of each action for each policy for a single patient over their ICU stay duration.  The discrete actions correspond to different combinations of vasopressor and IV fluid dosages. Agent 1 has a near-deterministic policy with an emphasis on a single treatment that is a combination of a low vasopressor dosage and a medium fluid dosage. Agents 2 and 3 are more aggressive than other agents, giving weights to medium and high vasopressor dosages as well. All agents emphasize taking action early rather than later in the duration (by which time hopefully the patient is in a stable state).  Importantly, all of these policies cannot be differentiated by value alone; thus they form a collection to be presented to clinicians for further review of potentially valuable options.  

%\textcolor{red}{Some space here to discuss the results -- exploration of diversity found. Perhaps also comparison with a simulator trained on the same data if time}

\section{Related Work}

Whereas we compute diversity (using MMD) between \emph{complete} trajectories (including state and action), many related works, quantify diversity only via differences in the action space. E.g., the KL diversity term in \citet{hong2018diversity} is between the actions. Similarly, in \citet{gangwani2018learning}, the kernel incorporates trajectory information but the definition of the target density is still according to the maximum entropy RL framework which focuses only on the diversity in action space. 

\citet{smith2018inference} learn a policy over options and can train multiple options (in an off-policy manner) using a rollout from a single option. The distinct options can give rise to behavior that is diverse, however there is no explicit diversity component in the objective and it is unclear how to summarize the kinds of distinct trajectories that are possible. Unlike \citet{eysenbach2018diversity}, our proposed method does not impose a latent structure defining `skills'.  \citet{cohen2019diverse} uses diversity for exploration in order to get closer to an optimal policy whereas our work solves multi-goal domains and identifies diverse policies of interest to experts (e.g. in clinical setting).%and what combination of options leads to the most interesting policies. 

In the graph-based planning literature (limited to the discrete setting), there is also an interest in finding diverse plans; \citet{srivastava2007domain} seek to do this using domain independent distance measures for evaluating diversity of plans whereas \citet{sohrabi2016finding} first generate a large set of high quality plans and then use clustering to identify a diverse set of representative plans that can be used for further analysis. Our motivation for seeking diverse policies in the reinforcement learning setting is aligned most closely with the end goal of \citet{sohrabi2016finding}: presenting a diverse set of representative solutions as a tool for hypothesis generation and to discover specific directions of interest for further inquiry.

There exists a literature on Bayesian methods for reinforcement learning \citep{ghavamzadeh2015bayesian}.  Also related are evolutionary \citep{lehman2011abandoning} and multi-objective \citep{liu2014multiobjective} reinforcement learning approaches. However, these approaches do not systematically attempt to identify a small set of distinct policies.

\section{Discussion}

Our DIPG algorithm successfully finds multiple goals in an optimal or near-optimal manner whereas baseline approaches are either unable to reach multiple goals or do so sub-optimally.  The poor task performance of SVPG could be due to the difficulties of performing functional gradient descent over a high-dimensional parameter space. However, even if those difficulties were to be overcome, the search for diversity in the space of neural network parameters does not correspond directly to any meaningful notion of diversity in the trajectories. The Soft-Q algorithm exhibits some meaningful diversity in the policies it finds (e.g. reaching different goals in the multi-goal environment), however, there is unnecessary stochasticity in the actions, leading to sub-optimal policies. Unlike the baselines, random restarts do result in near-optimal in-task performance.  However, the diversity of the policies is at the mercy of the basins of attraction of each local optima (based on the initialization and subsequent experience).  In contrast, DIPG-PPO successfully identifies a collection of distinct policies consistently and efficiently (that is, with few runs).    

%In the batch data setting, we are able to find policies that are of higher value than the behavior policy by using a surrogate objective that is easier to compute and optimize than the CWPDIS importance sampling estimator. Additionally, the surrogate encourages the policies to perform `safe' actions.  

While we have focused on certain notions of distinctness here (state visits, action counts), our approach extends easily to other measures of distinctness as measured by alternative kernels (see \citet{steinpoints2018} and \citet{gretton2012optimal} for novel kernels and discussion). Whereas the full trajectory may be needed for training the RL agent, we can capture (MMD-based) distinctness more generally using an arbitrary function of the trajectory. For example, we could group patients into clinically meaningful clusters which can be used to define a kernel for measuring diversity. There are also opportunities for incorporating more efficient search algorithms than gradient descent (e.g. \citet{17-toussaint-ICRA}).  

While we have focused on the task of returning policies as possible options to a human user, another use-case could be for situations in which we have a cheap, low fidelity simulator and a more expensive, high-fidelity simulator.  In this case, the distinct trajectories from the low-fidelity simulator could be used as seeds for deep exploration in the more expensive simulator. Diverse policies can help manage the exploration-exploitation trade-off \citep{bellemare2016unifying, fortunato2017noisy, fu2017ex2,cohen2018diverse}.

\section{Conclusion}
We presented an approach for identifying a collection of near-optimal
policies with significantly different distributions of trajectories.
Our MMD-based regularizer can be applied to the distribution of any
statistic of the trajectories---state visits, action counts,
state-action combinations---and can also be easily incorporated into
any policy-gradient method. We applied these to several standard
benchmarks and developed an off-policy extension to identify
meaningfully different treatment options from observational clinical
data. The ability to find these diverse policies may be useful when
the agent does not have complete information about the task, and for
presenting a set of potentially reasonable options to a downstream
human or system, who can use that information to efficiently choose
amongst reasonable options.

\section*{Acknowledgements}
We acknowledge support from AFOSR FA 9550-17-1-0155.

\bibliography{rl_bib}

\begin{thebibliography}{34}
\providecommand{\natexlab}[1]{#1}
\providecommand{\url}[1]{\texttt{#1}}
\expandafter\ifx\csname urlstyle\endcsname\relax
  \providecommand{\doi}[1]{doi: #1}\else
  \providecommand{\doi}{doi: \begingroup \urlstyle{rm}\Url}\fi

\bibitem[Bellemare et~al.(2016)Bellemare, Srinivasan, Ostrovski, Schaul,
  Saxton, and Munos]{bellemare2016unifying}
Marc Bellemare, Sriram Srinivasan, Georg Ostrovski, Tom Schaul, David Saxton,
  and Remi Munos.
\newblock Unifying count-based exploration and intrinsic motivation.
\newblock In \emph{Advances in Neural Information Processing Systems}, pages
  1471--1479, 2016.

\bibitem[{Chen} et~al.(2018){Chen}, {Mackey}, {Gorham}, {Briol}, and
  {Oates}]{steinpoints2018}
W.~Y. {Chen}, L.~{Mackey}, J.~{Gorham}, F.-X. {Briol}, and C.~J. {Oates}.
\newblock {Stein Points}.
\newblock \emph{ArXiv e-prints}, March 2018.

\bibitem[Cohen et~al.(2018)Cohen, Yu, and Wright]{cohen2018diverse}
Andrew Cohen, Lei Yu, and Robert Wright.
\newblock Diverse exploration for fast and safe policy improvement.
\newblock In \emph{Thirty-Second AAAI Conference on Artificial Intelligence},
  2018.

\bibitem[Cohen et~al.(2019)Cohen, Qiao, Yu, Way, and Tong]{cohen2019diverse}
Andrew Cohen, Xingye Qiao, Lei Yu, Elliot Way, and Xiangrong Tong.
\newblock Diverse exploration via conjugate policies for policy gradient
  methods.
\newblock \emph{arXiv preprint arXiv:1902.03633}, 2019.

\bibitem[Ellenberge(2019)]{PyBullet-Gym}
Benjamin Ellenberge.
\newblock pybullet-gym.
\newblock \url{https://github.com/benelot/pybullet-gym}, 2019.

\bibitem[Eysenbach et~al.(2018)Eysenbach, Gupta, Ibarz, and
  Levine]{eysenbach2018diversity}
Benjamin Eysenbach, Abhishek Gupta, Julian Ibarz, and Sergey Levine.
\newblock Diversity is all you need: Learning skills without a reward function.
\newblock \emph{arXiv preprint arXiv:1802.06070}, 2018.

\bibitem[Fard and Pineau(2011)]{fard2011non}
Mahdi~Milani Fard and Joelle Pineau.
\newblock Non-deterministic policies in markovian decision processes.
\newblock \emph{J. Artif. Intell. Res.(JAIR)}, 40:\penalty0 1--24, 2011.

\bibitem[Fortunato et~al.(2017)Fortunato, Azar, Piot, Menick, Osband, Graves,
  Mnih, Munos, Hassabis, Pietquin, et~al.]{fortunato2017noisy}
Meire Fortunato, Mohammad~Gheshlaghi Azar, Bilal Piot, Jacob Menick, Ian
  Osband, Alex Graves, Vlad Mnih, Remi Munos, Demis Hassabis, Olivier Pietquin,
  et~al.
\newblock Noisy networks for exploration.
\newblock \emph{arXiv preprint arXiv:1706.10295}, 2017.

\bibitem[Fu et~al.(2017)Fu, Co-Reyes, and Levine]{fu2017ex2}
Justin Fu, John Co-Reyes, and Sergey Levine.
\newblock Ex2: Exploration with exemplar models for deep reinforcement
  learning.
\newblock In \emph{Advances in Neural Information Processing Systems}, pages
  2577--2587, 2017.

\bibitem[Gangwani et~al.(2018)Gangwani, Liu, and Peng]{gangwani2018learning}
Tanmay Gangwani, Qiang Liu, and Jian Peng.
\newblock Learning self-imitating diverse policies.
\newblock \emph{arXiv preprint arXiv:1805.10309}, 2018.

\bibitem[Ghassemi et~al.(2017)Ghassemi, Wu, Hughes, Szolovits, and
  Doshi-Velez]{ghassemi2017predicting}
Marzyeh Ghassemi, Mike Wu, Michael~C Hughes, Peter Szolovits, and Finale
  Doshi-Velez.
\newblock Predicting intervention onset in the icu with switching state space
  models.
\newblock \emph{AMIA Summits on Translational Science Proceedings},
  2017:\penalty0 82, 2017.

\bibitem[Ghavamzadeh et~al.(2015)Ghavamzadeh, Mannor, Pineau, Tamar,
  et~al.]{ghavamzadeh2015bayesian}
Mohammad Ghavamzadeh, Shie Mannor, Joelle Pineau, Aviv Tamar, et~al.
\newblock Bayesian reinforcement learning: A survey.
\newblock \emph{Foundations and Trends{\textregistered} in Machine Learning},
  8\penalty0 (5-6):\penalty0 359--483, 2015.

\bibitem[Gretton et~al.(2007)Gretton, Borgwardt, Rasch, Sch{\"o}lkopf, and
  Smola]{gretton2007kernel}
Arthur Gretton, Karsten~M Borgwardt, Malte Rasch, Bernhard Sch{\"o}lkopf, and
  Alex~J Smola.
\newblock A kernel method for the two-sample-problem.
\newblock In \emph{Advances in neural information processing systems}, pages
  513--520, 2007.

\bibitem[Gretton et~al.(2012)Gretton, Sejdinovic, Strathmann, Balakrishnan,
  Pontil, Fukumizu, and Sriperumbudur]{gretton2012optimal}
Arthur Gretton, Dino Sejdinovic, Heiko Strathmann, Sivaraman Balakrishnan,
  Massimiliano Pontil, Kenji Fukumizu, and Bharath~K Sriperumbudur.
\newblock Optimal kernel choice for large-scale two-sample tests.
\newblock In \emph{Advances in neural information processing systems}, pages
  1205--1213, 2012.

\bibitem[Haarnoja et~al.(2017)Haarnoja, Tang, Abbeel, and
  Levine]{haarnoja2017reinforcement}
Tuomas Haarnoja, Haoran Tang, Pieter Abbeel, and Sergey Levine.
\newblock Reinforcement learning with deep energy-based policies.
\newblock \emph{arXiv preprint arXiv:1702.08165}, 2017.

\bibitem[Hong et~al.(2018)Hong, Shann, Su, Chang, Fu, and
  Lee]{hong2018diversity}
Zhang-Wei Hong, Tzu-Yun Shann, Shih-Yang Su, Yi-Hsiang Chang, Tsu-Jui Fu, and
  Chun-Yi Lee.
\newblock Diversity-driven exploration strategy for deep reinforcement
  learning.
\newblock In \emph{Advances in Neural Information Processing Systems}, pages
  10489--10500, 2018.

\bibitem[Johnson et~al.(2013)Johnson, Kramer, and Clifford]{johnson2013new}
Alistair E.~W. Johnson, Andrew~A Kramer, and Gari~D Clifford.
\newblock A new severity of illness scale using a subset of acute physiology
  and chronic health evaluation data elements shows comparable predictive
  accuracy.
\newblock \emph{Critical care medicine}, 41\penalty0 (7):\penalty0 1711--1718,
  2013.

\bibitem[Johnson et~al.(2017)Johnson, Stone, Celi, and
  Pollard]{johnson2017mimic}
Alistair~EW Johnson, David~J Stone, Leo~A Celi, and Tom~J Pollard.
\newblock The mimic code repository: enabling reproducibility in critical care
  research.
\newblock \emph{Journal of the American Medical Informatics Association},
  25\penalty0 (1):\penalty0 32--39, 2017.

\bibitem[Komorowski et~al.(2018)Komorowski, Celi, Badawi, Gordon, and
  Faisal]{komorowski2018artificial}
Matthieu Komorowski, Leo~A Celi, Omar Badawi, Anthony~C Gordon, and A~Aldo
  Faisal.
\newblock The artificial intelligence clinician learns optimal treatment
  strategies for sepsis in intensive care.
\newblock \emph{Nature Medicine}, 24\penalty0 (11):\penalty0 1716, 2018.

\bibitem[Lange et~al.(2012)Lange, Gabel, and Riedmiller]{lange2012batch}
Sascha Lange, Thomas Gabel, and Martin Riedmiller.
\newblock Batch reinforcement learning.
\newblock In \emph{Reinforcement learning}, pages 45--73. Springer, 2012.

\bibitem[Le~Gall et~al.(1993)Le~Gall, Lemeshow, and Saulnier]{le1993new}
Jean-Roger Le~Gall, Stanley Lemeshow, and Fabienne Saulnier.
\newblock A new simplified acute physiology score (saps ii) based on a
  european/north american multicenter study.
\newblock \emph{Jama}, 270\penalty0 (24):\penalty0 2957--2963, 1993.

\bibitem[Lehman and Stanley(2011)]{lehman2011abandoning}
Joel Lehman and Kenneth~O Stanley.
\newblock Abandoning objectives: Evolution through the search for novelty
  alone.
\newblock \emph{Evolutionary computation}, 19\penalty0 (2):\penalty0 189--223,
  2011.

\bibitem[Levine and Koltun(2013)]{levine2013guided}
Sergey Levine and Vladlen Koltun.
\newblock Guided policy search.
\newblock In \emph{International Conference on Machine Learning}, pages 1--9,
  2013.

\bibitem[Liu et~al.(2014)Liu, Xu, and Hu]{liu2014multiobjective}
Chunming Liu, Xin Xu, and Dewen Hu.
\newblock Multiobjective reinforcement learning: A comprehensive overview.
\newblock \emph{IEEE Transactions on Systems, Man, and Cybernetics: Systems},
  45\penalty0 (3):\penalty0 385--398, 2014.

\bibitem[Liu et~al.(2017)Liu, Ramachandran, Liu, and Peng]{liu2017stein}
Yang Liu, Prajit Ramachandran, Qiang Liu, and Jian Peng.
\newblock Stein variational policy gradient.
\newblock \emph{arXiv preprint arXiv:1704.02399}, 2017.

\bibitem[Schulman et~al.(2017)Schulman, Wolski, Dhariwal, Radford, and
  Klimov]{schulman2017proximal}
John Schulman, Filip Wolski, Prafulla Dhariwal, Alec Radford, and Oleg Klimov.
\newblock Proximal policy optimization algorithms.
\newblock \emph{arXiv preprint arXiv:1707.06347}, 2017.

\bibitem[Smith et~al.(2018)Smith, Hoof, and Pineau]{smith2018inference}
Matthew Smith, Herke Hoof, and Joelle Pineau.
\newblock An inference-based policy gradient method for learning options.
\newblock In \emph{International Conference on Machine Learning}, pages
  4710--4719, 2018.

\bibitem[Sohrabi et~al.(2016)Sohrabi, Riabov, Udrea, and
  Hassanzadeh]{sohrabi2016finding}
Shirin Sohrabi, Anton~V Riabov, Octavian Udrea, and Oktie Hassanzadeh.
\newblock Finding diverse high-quality plans for hypothesis generation.
\newblock In \emph{ECAI}, pages 1581--1582, 2016.

\bibitem[Sriperumbudur et~al.(2010)Sriperumbudur, Fukumizu, Gretton,
  Sch{\"o}lkopf, and Lanckriet]{sriperumbudur2010non}
Bharath~K Sriperumbudur, Kenji Fukumizu, Arthur Gretton, Bernhard
  Sch{\"o}lkopf, and Gert~RG Lanckriet.
\newblock Non-parametric estimation of integral probability metrics.
\newblock In \emph{Information Theory Proceedings (ISIT), 2010 IEEE
  International Symposium on}, pages 1428--1432. IEEE, 2010.

\bibitem[Srivastava et~al.(2007)Srivastava, Nguyen, Gerevini, Kambhampati, Do,
  and Serina]{srivastava2007domain}
Biplav Srivastava, Tuan~Anh Nguyen, Alfonso Gerevini, Subbarao Kambhampati,
  Minh~Binh Do, and Ivan Serina.
\newblock Domain independent approaches for finding diverse plans.
\newblock In \emph{IJCAI}, pages 2016--2022, 2007.

\bibitem[Sutton et~al.(2000)Sutton, McAllester, Singh, and
  Mansour]{sutton2000policy}
Richard~S Sutton, David~A McAllester, Satinder~P Singh, and Yishay Mansour.
\newblock Policy gradient methods for reinforcement learning with function
  approximation.
\newblock In \emph{Advances in neural information processing systems}, pages
  1057--1063, 2000.

\bibitem[Thomas(2015)]{thomas2015safe}
Philip~S Thomas.
\newblock \emph{Safe reinforcement learning}.
\newblock PhD thesis, University of Massachusetts Libraries, 2015.

\bibitem[Toussaint and Lopes(2017)]{17-toussaint-ICRA}
Marc Toussaint and Manuel Lopes.
\newblock Multi-bound tree search for logic-geometric programming in
  cooperative manipulation domains.
\newblock In \emph{(ICRA 2017)}, 2017.

\bibitem[Vincent et~al.(1996)Vincent, Moreno, Takala, Willatts,
  De~Mendon{\c{c}}a, Bruining, Reinhart, Suter, and Thijs]{vincent1996sofa}
J-L Vincent, Rui Moreno, Jukka Takala, Sheila Willatts, Arnaldo
  De~Mendon{\c{c}}a, Hajo Bruining, CK~Reinhart, PeterM Suter, and LG~Thijs.
\newblock The sofa (sepsis-related organ failure assessment) score to describe
  organ dysfunction/failure.
\newblock \emph{Intensive care medicine}, 22\penalty0 (7):\penalty0 707--710,
  1996.

\end{thebibliography}
\bibliographystyle{plainnat}

\clearpage
\appendix{}
\section{Additional Experiments}

\subsection{On-Policy DIPG-PPO}

\begin{figure}[h]
\begin{multicols}{2}

    \includegraphics[width=\linewidth]{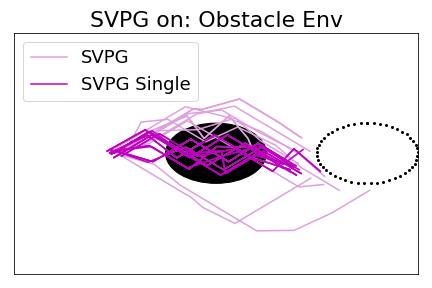}\par 
    \includegraphics[width=\linewidth]{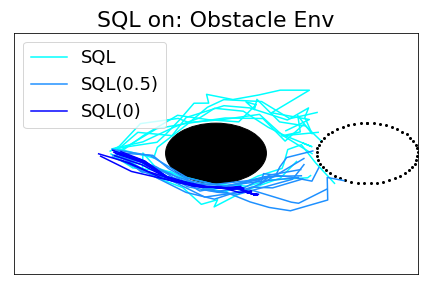}\par 
    \includegraphics[width=\linewidth]{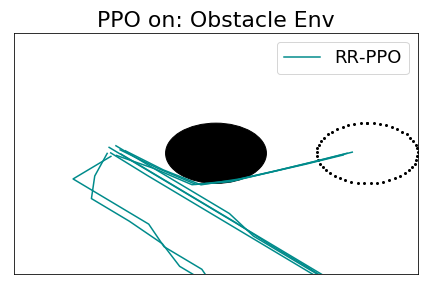}\par
    \includegraphics[width=\linewidth]{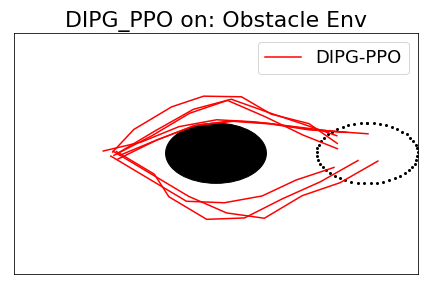}\par  
    \end{multicols}
\caption{SVPG fails to avoid the obstacle in most cases. Soft-Q (under default entropy regularization) learns sub-optimal diverse paths to successfully avoid the obstacle, Soft-Q with smaller and no entropy regularization finds only a single path around the obstacle. RR-PPO only successfully finds one path (in an optimal manner) however in many instances it is unable to find the goal. Perhaps there is a need for many more restarts. DIPG-PPO successfully finds both paths to the goal.}
\label{fig:barrier_trajs}
\end{figure}

We investigated the performance of our algorithm and compared to baseline algorithms on another 2-D navigational domain. The results in figure~\ref{fig:barrier_trajs} how how successfully DIPG-PPO solves this domain in comparison with other baseline algorithms.

\textit{Obstacle} environment involves overcoming a pit/barrier region to reach the goal. There exist two optimal policies (going around the barrier in two different ways) that are meaningfully distinct. This environment requires finding different policies to reach a single goal. 

\subsection{Off-Policy Experiments}

\begin{figure}[h]
    \includegraphics[width=0.9\linewidth]{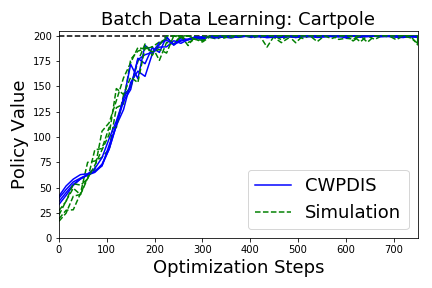}\par  
\caption{We show the learning curve for cartpole batch data under the CWPDIS estimator as well as through actual simulation of the policy. The ability of our surrogate-based objective to generalize from a small batch of trajectories is demonstrated by the consistently high returns achieved. We simulate using a trajectory length of 200, so the highest possible policy value is 200.}
\label{fig:batch_cartpole}
\end{figure}

Batch data from \textit{Cartpole} (270 trajectories) was used for training using the proposed surrogate in the paper. We find that this surrogate is consistently able to solve the cartpole domain as evaluated on `ground truth' simulation (trajectory length = 200) of the policies as well as the CWPDIS estimator (figure~\ref{fig:batch_cartpole}).  

\section{Hypotension Cohort}

Our data source is the publicly-available MIMIC-III database \cite[]{johnson2016mimic}.  The full dataset contains static and dynamic information for nearly 60,000 patients treated in the critical care units of Beth-Israel Deaconess Medical Center (BIDMC) in Boston between 2001-2012. In our work we use version 1.4 of MIMIC-III, released in September 2016. Much of our data processing is reused from original queries from \cite{ghassemi2017predicting}, with some additional processing pulled from the public code released by \cite{komorowski2018artificial}. In Table \ref{tab:baseline}, we present some baseline characteristics of the selected cohort.

\begin{table*}[]
\begin{tabular}{l | l}
Characteristic                                 & Overall Cohort (N=9,860) \\ \hline
Age (25/50/75\% quantiles)                              & 57.4, 69.4, 80.5         \\
Female (\%)                                             & 47.6\%                   \\
Weight in kg (25/50/75\% quantiles)                     & 65.0, 77.1, 90.9         \\
Surgical ICU (\%)                                       & 48.2\%                   \\
Initial SOFA score                                      & 2, 4, 7                  \\
Initial OASIS score                                     & 27, 33, 39               \\
Initial SAPS score                                      & 15, 19, 22               \\
Non-white (\%)                                          & 23.7\%                   \\
Emergency Admission                                     & 82.4\%                   \\
Urgent Admission                                        & 1.3\%                    \\
Time from Admission to ICU (mean; 25/50/75\% quantiles) & 48.5; 0.0, 0.1, 26.9     
\end{tabular}
\caption{Baseline characteristics of the set of ICU stays used in our experiments.}
\label{tab:baseline}
\end{table*}

\subsection{Cohort Selection} 
Each observation in our final dataset is a single ICU stay. In some cases it is possible that a single patient appears multiple times if they had multiple ICU stays within a single hospital admission, or if they had multiple admissions to BIDMC during the period of interest.

We started with all non-pediatric patients, filtering to only those with an age of at least 15 on admission. We further filtered to only include patients in MIMIC-III that came from MetaVision, as these were the only patients where we could reliably and easily extract both start and end times for the interventions of interest. Next, we filtered out the bottom and top 1\% quantiles in terms of length of stay, so that only ICU stays that were at least 14 hours but no longer than 622 hours were included. Lastly, we filtered to only include ICU stays that had at least three distinct measurements of mean arterial pressure (MAP), and at least one MAP below 65, indicating hypotension.  This resulted in a cohort of 9,860 ICU stays.

\subsection{Data Extraction} 

Our dataset contains 11 static features available on admission to the ICU (or shortly thereafter): age, biological sex, weight, whether the ICU was a surgical ICU, three overall severity scores (SOFA \cite[]{vincent1996sofa}, OASIS \cite[]{johnson2013new}, and SAPS \cite[]{le1993new}), race, whether the hospital admission was urgent, whether the hospital admission was an emergency, and hours from hospital admission to ICU admission. 

We also have a total of 20 clinical time series variables measured over the course of a patient's ICU stay. These include 8 vitals: diastolic blood pressure, heart rate, mean arterial pressure (MAP), systolic blood pressure, pulse oximetry, respiration rate, temperature, and urine output. Vitals are typically recorded about once an hour, although in practice they are captured at the bedside continuously.   We also include 12 laboratory measurements: fraction of inspired oxygen, blood urea nitrogen, creatinine, glucose, bicarbonate, hematocrit, lactate, magnesium, platelets, potassium, sodium, and white blood cell count. These are typically only measured a few times a day from blood samples drawn from patients. 

Lastly, we extracted information on the interventions of interest: fluid therapy and vasopressor therapy. We combine different types of fluids and blood products together when forming our fluid action variable; in particular, we filter to only include common NaCl 0.9\% solution, lactated ringer's, packed red blood cells, fresh frozen plasma, and platelets. We include five different types of vasopressors for the vasopressor action variable: dopamine, epinephrine, norepinephrine, vasopressin, and phenylephrine. We map these five drugs into a common dosage amount based off norepinephrine equivalents, following the preprocessing in \cite{komorowski2018artificial}.
% FDV: Overall this is great detail and well organized!  One thing the clinicians asked: is it dose or dose per kg weight patient?  

\subsection{Feature Choices} 

The state space in our RL formulation consists of the 31 previously listed variables. We discretize time into 30 minute windows and impute any unobserved vital or lab value with the population median value. Once a variable is observed in a given hospital admission, we then use the last observed measurement of a variable until a new value is measured. 

%In addition, for the labs, we include binary indicator variables that denote whether a lab was measured since the last time, as labs are typically informatively missing and the clinical decision to measure a certain lab can be important \cite[]{agniel2018biases}.  

We discretize the two types of interventions into six different dosage levels (including zero dosage) and allow for any combination of the two dosage levels to exist as an action. In total, this results in 36 unique actions that may be taken at each decision time.

\end{document}